\documentclass{article}
\usepackage{spconf,amsmath,graphicx}
\usepackage{hyperref}
\usepackage{multirow}
\usepackage{makecell}
\usepackage{float}

\title{Generating Synthetic Audio Data for Attention-Based\\ Speech Recognition Systems}
\name{Nick Rossenbach, Albert Zeyer, Ralf Schl\"uter, Hermann Ney}
\address{Human Language Technology and Pattern Recognition, Computer Science Department,\\
RWTH Aachen University, 52074 Aachen, Germany\\
AppTek GmbH, 52062 Aachen, Germany\\
{\tt \textless surname\textgreater@i6.informatik.rwth-aachen.de}\vspace{-0.25em}}
\usepackage[tracking=true]{microtype}
\usepackage{wrapfig}

\makeatletter
\renewcommand{\section}{\@startsection
  {section}%
  {1}%
  {}%
  {-1.0\baselineskip}%
  {0.6\baselineskip}%
  {}}%

\renewcommand{\subsection}{\@startsection
  {subsection}%
  {2}%
  {}%
  {-0.8\baselineskip}%
  {0.4\baselineskip}%
  {}}%
  
\usepackage{caption}
\newcommand\T{\rule{0pt}{2.1ex}}       

\usepackage{cite}

\begin{document}
%
\maketitle
\begin{abstract}
Recent advances in text-to-speech (TTS) led to the development of flexible multi-speaker end-to-end TTS systems. We extend state-of-the-art attention-based automatic speech recognition (ASR) systems with synthetic audio generated by a TTS system trained only on the ASR corpora itself. ASR and TTS systems are built separately to show that text-only data can be used to enhance existing end-to-end ASR systems without the necessity of parameter or architecture changes. We compare our method with language model integration of the same text data and with simple data augmentation methods like SpecAugment and show that performance improvements are mostly independent. We achieve improvements of up to 33\% relative in word-error-rate (WER) over a strong baseline with data-augmentation in a low-resource environment (LibriSpeech-100h), closing the gap to a comparable oracle experiment by more than 50\%. We also show improvements of up to 5\% relative WER over our most recent ASR baseline on LibriSpeech-960h. 
\end{abstract}
\section{Introduction \& Related Work}
\label{sec:intro}

Recently published automatic speech recognition (ASR) systems are based on deep neural network approaches, either in combination with hidden-markov-models (hybrid approach) or as a standalone end-to-end system. While hybrid deep neural network architectures provide state-of-the-art performance, recent results using end-to-end architectures show competing performance on large resource tasks \cite{DBLP:journals/corr/abs-1905-03072}. Improvements were achieved by using new data augmentation methods such as SpecAugment \cite{DBLP:journals/corr/abs-1904-08779} or using advanced pre-training schemes \cite{DBLP:conf/interspeech/ZeyerISN18}. For medium to low resource tasks, hybrid architectures are still superior to end-to-end approaches \cite{DBLP:journals/corr/abs-1905-03072}. To further increase the performance of end-to-end systems in low resource conditions, untranscribed speech or text can be used as additional training data. A previously published approach is the text-to-encoder (TTE) model which can integrate additional text \cite{DBLP:conf/slt/HayashiWZTHAT18} or untranscribed speech \cite{DBLP:journals/corr/abs-1811-01690} into ASR training. Another method is the joint training of ASR and text-to-speech (TTS) systems such as the Speech Chain approach \cite{DBLP:conf/asru/TjandraS017, DBLP:conf/interspeech/TjandraS018, DBLP:conf/icassp/TjandraS019} or variants of it \cite{DBLP:journals/corr/abs-1905-01152}. Training TTS systems on external data to create audio features for ASR has also been investigated in \cite{DBLP:conf/slt/MimuraUISK18, DBLP:journals/corr/abs-1811-00707}. The usage of TTS in the context of ASR training was inspired by recent advances in end-to-end TTS systems with multispeaker capabilities such as Tacotron \cite{Wang2017}, Tacotron-2 \cite{DBLP:journals/corr/abs-1712-05884} and Deep-Voice \cite{DBLP:journals/corr/abs-1710-07654}.

Most of the previously presented approaches require that the ASR and TTS systems share at least a common feature processing pipeline and operate on the same kind of audio features. Especially for approaches where ASR and TTS are trained jointly, this is a strict requirement. In contrast to that, our approach includes a completely separate end-to-end TTS system with a Griffin \& Lim (G\&L) vocoder \cite{DBLP:conf/icassp/GriffinDL84} for synthetic waveform generation instead of synthetic feature generation. While related work covering independent TTS systems \cite{DBLP:conf/slt/MimuraUISK18, DBLP:journals/corr/abs-1811-00707} uses additional data, our TTS system is only trained on the ASR corpus itself. The synthetic data is stored as compressed audio and can be used for any kind of speech recognition system, with no relation to the TTS system. For adaptive speaker embeddings, we compare global-style-tokens (GST) \cite{DBLP:conf/icml/WangSZRBSXJRS18} and i-vector representations \cite{DBLP:journals/corr/abs-1906-06207}. To the best of our knowledge, previous work on integrating synthesized data from TTS systems to ASR did not include a comparison with other data-augmentation techniques. Thus, we compare our TTS approach to SpecAugment \cite{DBLP:journals/corr/abs-1904-08779} and speed-perturbation \cite{conf/interspeech/KoPPK15}. We also include a direct comparison with language-model integration of the same text data as used to generate synthetic speech. The experiments in this work were managed with Sisyphus \cite{DBLP:conf/emnlp/PeterBN18} and the ASR and TTS systems were implemented in RETURNN \cite{DBLP:conf/icassp/DoetschZVKSN17}. The RETURNN configs will be made publicly available\footnote{\scriptsize\url{https://github.com/rwth-i6/returnn-experiments/tree/master/2019-asr-synthetic-data}}.


\section{Attention-based Speech Recognition}
\label{sec:format}

\subsection{Model}
Our baseline model for \textit{LibriSpeech-100h} consists of 6 bi-directional long-short-term-memory (LSTM) encoder layers, and a single LSTM decoder layer, following \cite{DBLP:conf/interspeech/ZeyerISN18}. The encoder layers have a dimension of 1024 each. The decoder layer has a dimension of 1000. The time dimension is reduced by max-pooling with a factor of 2 between the first three layers, resulting in a total time reduction of 8. The attention mechanism is a MLP-style attention with weight feedback \cite{DBLP:conf/interspeech/ZeyerISN18}. The dimension of the attention-MLP combination is 1024. 
CTC loss \cite{Graves:2006:CTC:1143844.1143891} is used as additional training criterion besides cross-entropy (CE), based on label prediction with a softmax layer taking the encoder states as input.

As input we use 40-dimensional MFCC features. Each feature dimension is normalized to zero mean and variance of one by estimating statistics on the training data. The training includes a pre-training scheme as in \cite{DBLP:conf/interspeech/ZeyerISN18}. For the text labels we use byte-pair-encoding \cite{DBLP:conf/acl/SennrichHB16a} with 10k merge operations.
%
%

\subsection{Data Augmentation}

For each experiment we include a variant of SpecAugment \cite{DBLP:journals/corr/abs-1904-08779} as data augmentation method. The spectral axes of the features are masked randomly at 1 to 4 positions spanning between 1 and 8 features. On the time axis, we mask between 1 and $\frac{1}{50}$ of the number of frames positions, for a maximum of 20 consecutive frames. We also compare SpecAugment to speed-perturbation \cite{conf/interspeech/KoPPK15}, using the perturbation factors 0.9, 0.95, 1.05 and 1.1 to add 4 times the original data.

\subsection{Language Model}

For some experiments we use two additional language models trained on either the text data used for synthesis (\textit{LM-small}) or all of the additional data for language modeling (\textit{LM-large}). The language models are based on the transformer architecture, similar to work presented in \cite{DBLP:journals/corr/abs-1905-04226}. \textit{LM-small} is a 30 layer tranformer architecture with a feed-forward dimension of 2048 and an attention dimension of 512. For \textit{LM-large} we use a 32 layer transformer architecture with a feed-forward dimension of 4096 and an attention dimension of 1024. The language model is added to the ASR system with log-linear combination. 
%

\section{Speech Synthesis System}

\subsection{Synthesis Network}
The synthesis network is inspired by the Tacotron-2 architecture \cite{DBLP:journals/corr/abs-1712-05884}, but contains some modifications. We use 80 dimensional log-mel features with a preemphasis factor of $\alpha = 0.97$, a window size of $50ms$ and a window shift of $12.5ms$. The features are globally normalized to zero mean and variance of one by extracting feature statistics from the training data. We set a bound to the mel-scale at 60 Hz, removing lower frequency bands. The input symbols are lowercased characters, and we add an additional end-of-sequence token "$\sim$".

The encoder consists of 3 1-D convolutional layers with 128 filters of size 5, and a bi-directional LSTM with 128 hidden units each forming the encoder states $h_j$ for a character sequence $c_1^J$.
The attention mechanism is an MLP-attention with convolutional weight feedback \cite{DBLP:journals/corr/ChorowskiBSCB15} using the sum of attention weights. The decoder consists of two stacked LSTM layers, from which we only use the second layer $s_i^{(2)}$  as input to the attention. In addition to the encoder state, a 64 dimensional positional encoding is used in the attention. The attention energy is computed as:
\begin{equation}
e_{i,j} = v^T \tanh(W_s s_i^{(2)} + W_h h_j + W_p \text{posenc(j)} + W_\gamma \gamma_i)
\end{equation}

The convolutional weight feedback $\gamma_i$ is computed with 32 1-D convolutional filters of size 31, applied on the sum of previous alignments. Instead of performing zero padding, we pad the positions before the first encoder state with ones, to indicate that positions before the start are already "attended".

The output is a single linear layer, transforming the decoder state and the current context vector into the shape of the output features. We stack 3 frames per single output step to reduce the sequence length and increase the attention stability. The input for the stop token is a linear transformation of the same inputs, predicting a single scalar with an applied sigmoid that indicates a finished sequence.
We use L1 loss for the spectral features and binary cross-entropy loss for the stop token. The stop token target is a "ramp" of length 5, meaning that the target values for the binary CE loss are $[0,...,0,0.2,0.4,0.6,0.8,1.0]$. To prohibit early stopping during decoding, we continue for 5 decoder steps (15 frames) after the stop token value exceeds a threshold of $0.4$.

\subsection{Speaker Modeling}

We use two different methods to enable speaker adaptation in the TTS system. One is a GST \cite{DBLP:conf/icml/WangSZRBSXJRS18} based embedding, which is an unsupervised method of adaption. The other uses i-vector representations computed as in \cite{DBLP:journals/corr/abs-1906-06207}. 

For the GST speaker embedding we use 6 2-D convolutional layers with stride 2 to extract a short feature sequence from the target audio. A single feature vector is computed by applying a single forward LSTM on the sequence and taking the last state. This feature vector is used to select a mixture from the style token feature bank containing 100 entries of size 128 via attention. The mixture of style tokens or an i-vector representation is concatenated to the LSTM encoder states to form the speaker adapted encoder representation.

\subsection{Data Preprocessing}

We found that the audio data of the LibriSpeech corpus is designed in a way that is not beneficial for TTS systems. The speech utterances are not based on full sentences, meaning that there can be 2 sentences in one utterance, with style changes or unnatural pauses in between. Some utterances start or end in the middle of a sentence, leading to unnatural pronounciation at the beginning and end of utterances. These problems were also adressed in \cite{libritts}. To remove unnatural pauses and long pauses in general, we apply the FFMPEG silenceremove filter\footnote{\scriptsize\url{https://ffmpeg.org/ffmpeg-filters.html\#silenceremove}} with a threshold of -40dB.

\subsection{Generation}

We train a seperate network that converts the log-mel features into linear features, which are necessary for direct G\&L conversion. In the Tacotron architecture \cite{Wang2017}, the network converting log-mel features to linear spectograms is part of the same training. We chose to train the two parts separately, so the linear network needs to be trained only once on the available audio data.

%
%

\begin{table}[t]
\caption{Results on LibriSpeech-100 with SpecAugment and Speed Perturbation.}
\label{results100aug}
\begin{center}
\begin{tabular}{|c|c|c|c|c|c|}
\hline
\multirow{3}{*}{\makecell{Spec \\ Aug}} & \multirow{3}{*}{\makecell{Speed \\ Pert.}} & \multicolumn{4}{c|}{WER[\%]}\T\\
\cline{3-6}
& &  \multicolumn{2}{c|}{dev} & \multicolumn{2}{c|}{test}\T\\
\cline{3-6}
& & clean & other & clean & other\T\\
\hline
\hline
\multirow{2}{*}{No} & No & 12.8 & 36.8 & 12.8 & 38.7\T\\
\cline{2-6}
& Yes & 11.8 & 34.6 & 11.8 & 37.1\T\\
\cline{1-6} 
\multirow{2}{*}{Yes} & No & \textbf{10.5} & 27.7 & \textbf{10.8} & \textbf{28.8}\T\\
\cline{2-6}
& Yes & 10.6 & \textbf{27.2} & 10.9 & \textbf{28.8}\T\\
\hline
\end{tabular}
\end{center}
\vspace{-2.0em}
\end{table}

The mel-to-linear network consists of 2 stack BLSTM layers with a residual connections. The outputs are 512-dimensional linear spectograms (the DC-part is excluded). 
When generating the audio data we run the feature network on the text data and apply the mel-to-linear network on the resulting features. The linear features are used as input to a G\&L vocoder. We only use a single iteration for phase reconstruction as there is no need to reconstruct the phase when using MFCC features in the ASR system. 
Because both neural models perform regression tasks without search, and G\&L synthesis with a single iteration is computationally cheap, we can generate large amounts of training data in a short time period. Generating 30,000 utterances with 50 hours of speech takes 60 minutes for the feature generation, 10 minutes for the conversion and 60 minutes for G\&L synthesis and file encoding using a machine with 4 CPU cores and a single GPU.

\section{Experiments}

\begin{table}[!t]

\caption{Results on LibriSpeech-100 comparing synthetic audio data against LM-combination based on the same additional text data (\textit{LM-small}) and SpecAugment. We compare the results of our paper (*) with the results presented in \cite{DBLP:journals/corr/abs-1905-01152}. The model CCT$^{\dagger}$ also uses untranscribed speech.} 
\label{results100lm}
\begin{center}
\begin{tabular}{|c|c|c|c|c|c|c|c|}
\hline
\multirow{3}{*}{P} & \multirow{3}{*}{\makecell{Spec \\ Aug}} & \multirow{3}{*}{\makecell{Syn. \\ Data}} &\multirow{3}{*}{LM} & \multicolumn{4}{c|}{WER[\%]}\T\\
\cline{5-8}
& & & &  \multicolumn{2}{c|}{dev} & \multicolumn{2}{c|}{test}\T\\
\cline{5-8}
& & & & cl. & oth. & cl. & oth.\T\\
\hline
\hline
\multirow{10}{*}{*} & \multirow{5}{*}{No} & \multirow{2}{*}{No} & N &12.8 & 36.8 & 12.8 & 38.7\T\\
\cline{4-8}
& & & Y & 11.4 & 34.5 & 11.4 & 36.4\T\\
\cline{3-8}
& & \multirow{2}{*}{GST} & N & 10.2 & 34.9 & 10.6 & 36.9\T\\
\cline{4-8}
& & & \multirow{2}{*}{Y} & \phantom{0}8.9 & 33.0 & \phantom{0}9.3 & 34.8\T\\
\cline{5-8}
\cline{3-3}
& & oracle & & \phantom{0}5.9 & 22.6 & \phantom{0}6.2 & 23.5\T\\
\cline{2-8}
& \multirow{5}{*}{Yes} & \multirow{2}{*}{No} & N &10.5 & 27.7 & 10.8 & 28.8\T\\
\cline{4-8}
& & & Y & \phantom{0}9.9 & 27.0 & 10.3 & 28.1\T\\
\cline{3-8}
& & \multirow{2}{*}{GST} & N & \phantom{0}8.2 & 27.4 & \phantom{0}8.7 & 28.4\T\\
\cline{4-8}
& & & \multirow{2}{*}{Y} & \phantom{0}\textbf{7.4} & \textbf{25.7} & \phantom{0}\textbf{7.9} & \textbf{26.7}\T\\
\cline{3-3}
\cline{5-8}
& & oracle & & \phantom{0}5.4 & 17.9 & \phantom{0}5.6 & 18.4\T\\
\hline
\hline
\multirow{5}{*}{\cite{DBLP:journals/corr/abs-1905-01152}} & \multirow{5}{*}{No} & - & \multirow{2}{*}{N} & & & 21.9 & \T\\ 
\cline{5-8}
\cline{3-3}
& & \multirow{2}{*}{x-vec} & & & & 17.9 & \T\\ 
\cline{4-8}
& & & \multirow{2}{*}{Y} & & & 17.0 & \T\\ 
\cline{5-8}
\cline{3-3}
& & +CCT$^{\dagger}$& & & & 16.6 & \T\\
\cline{3-4}
\cline{5-8}
& & oracle & N & & & 11.8 & \T\\
\hline
\end{tabular}
\end{center}
\vspace{-2.5em}
\end{table}
We performed our simulated low-resource experiments on LibriSpeech \cite{LibriSpeech}, similar to previous work (e.g. \cite{DBLP:journals/corr/abs-1905-01152}). We use \textit{LibriSpeech-100h} as training data for the ASR baseline and the TTS system and the transcriptions of \textit{LibriSpeech-360h} as text-only data. We trained the baseline ASR system for 80 checkpoints and 4 data epochs to reach an initially converged state. This reduces the variance in the resulting performance, as all comparable training runs start from the same checkpoint. We then reset the learning rate and continue the training for an additional 170 checkpoints including data-augmentation methods and/or synthetic data. We set the epoch partitioning factor for each checkpoint in a way that for each experiment roughly the same amount of audio data was seen during training, independent from the amount of synthetic data generated. The parameter for LM score combination is optimized on dev-clean for test-clean, and on dev-other for test-other.

\subsection{Results for LibriSpeech-100}

Before adding synthetic data, we compared the effects of speed-perturbation and SpecAugment. The results can be seen in Table \ref{results100aug}. In our setting, SpecAugment and speed-perturbation both show improvements over the baseline, but the improvement of speed-perturbation vanishes when being combined with SpecAugment. For the following experiments we only use SpecAugment as data-augmentation method. In a next step, we used the text of the \textit{LibriSpeech-360h} corpus to generate synthetic audio with the GST-TTS model (Table \ref{results100lm}). To be able to compare the effect of additional audio to the effect of additional text, we included the \textit{LM-small} language model. In direct comparsion, adding synthetic data showed a better performance on test-clean, while adding an LM showed better performance on test-other. By combining both, we achieved a relative improvement of 27\% over the baseline on test-clean. The same relative improvement was observed over a stronger baseline including SpecAugment. The orcale experiment includes the text and audio of \textit{LibriSpeech-360h}, but uses the same initial checkpoint as all other experiments (only \textit{LibriSpeech-100h} for the first 4 epochs) to be comparable. While on test-clean we can reduce the gap to the oracle performance by more than 50\%, we only see a small improvement on test-other. We compare our results with the x-vector (x-vec) TTS system and the cycle-consitent (joint) training (CCT) of ASR and TTS models presented in \cite{DBLP:journals/corr/abs-1905-01152}. With their joint model, they achieved a relative improvement of 21\% on test-clean over a weaker baseline. Scores on test-other were not reported.

\begin{table}[!t]
\caption{Results on LibriSpeech-100 comparing synthetic data generated with a TTS system using GST or i-vector embeddings against the oracle data. All results are with an additional LM (\textit{LM-large}).}
\label{results100syn}
\begin{center}
\begin{tabular}{|c|c|c|c|c|c|}
\hline
\multirow{3}{*}{\makecell{Spec \\ Aug}} & \multirow{3}{*}{\makecell{Syn. \\ Audio}} & \multicolumn{4}{c|}{WER[\%]}\T\\
\cline{3-6}
& & \multicolumn{2}{c|}{dev} & \multicolumn{2}{c|}{test}\T\\
\cline{3-6}
& & clean & other & clean & other\T\\
\hline
\hline
\multirow{4}{*}{No} & - & 8.5 & 30.7 & 8.8 & 32.5\T\\
\cline{2-6}
& GST & 6.1 & 29.7 & 6.5 & 30.8\T\\
\cline{2-6}
& i-vector & 7.2 & 30.8 & 7.4 & 32.7\T\\
\cline{2-6}
& oracle & 3.9 & 18.7 & 4.2 & 19.2\T\\
\hline
\hline
\multirow{4}{*}{Yes} & - & 7.3 & 23.3 & 8.1 & 24.5\T\\
\cline{2-6}
& GST & \textbf{5.0} & \textbf{21.7} & \textbf{5.4} & \textbf{22.2}\T\\
\cline{2-6}
& i-vector& 5.6 & 23.5 & 6.0 & 24.5\T\\
\cline{2-6}
& oracle & 3.7 & 15.4 & 4.2 & 15.7\T\\
\hline
\end{tabular}
\end{center}
\vspace{-1.2em}
\end{table}

In Table \ref{results100syn} we compare the use of synthetic data generated with two different speaker embedding methods against an oracle experiment. The GST-based system clearly outperforms the TTS system using i-vector representations.  We see that the relative improvement of using synthetic data gets even larger when using \textit{LM-large} and SpecAugment, up to a performance increase of 33\%. The amount of presented audio during training is about 4000 hours, corresponding to $\sim$12.5 epochs original data (100h) and 8.5 epochs synthetic data (330h).

%
%

\subsection{Results for LibriSpeech-960}

In preliminary experiments we also tested if we can improve our currently best baseline setups for \textit{LibriSpeech-960h} using the TTS models trained on \textit{LibriSpeech-100h}. For the i-vector model we used representations computed on the full corpus, but the GST model is exactly the same as in the 100h case. We generated 2000 hours of additional data with the TTS models using the language model text-data as input, and used the on-the-fly feature augmentation to train the models for 12,500 hours of augmented data. The results for using synthetic data from TTS together with our best baseline using SpecAugment and \textit{LM-large} can be found in Table \ref{results1000syn}. As the models converged slower and showed less overfitting when using synthetic data, we used an original to synthetic data ratio of 3 to 2 and extended the training time by re-training each model for another 12,500 hours of training data ($\sim$15 epochs on the original data and $\sim$5 epochs on the synthetic data). The relative improvements are not exceeding 5\% WER, and we assume further investigation is needed to balance the regularization effects of Dropout, SpecAugment and Synthetic Data. We compare our results to the improvement achieved by training a separate TTS on the 3 Speaker M-AILABS corpus as performed in \cite{DBLP:journals/corr/abs-1811-00707}.

\begin{table}[!t]

\caption{Results on LibriSpeech-960 comparing synthetic data generated with a TTS system using GST or i-vector embeddings against the oracle data. In contrast to \cite{DBLP:journals/corr/abs-1811-00707}, our results include SpecAugment and an LM (\textit{LM-large}) in decoding.}
\label{results1000syn}
\begin{center}
\begin{tabular}{|c|c|c|c|c|c|c|}
\hline
\multirow{3}{*}{Paper} &  \multirow{3}{*}{Retrain} & \multirow{3}{*}{\makecell{Syn. \\ Audio}} & \multicolumn{4}{c|}{WER[\%]}\\
\cline{4-7}
& & & \multicolumn{2}{c|}{dev} & \multicolumn{2}{c|}{test}\\
\cline{4-7}
& & & cl. & oth. & cl. & oth.\T\\
\hline
\hline
\multirow{6}{*}{\textit{\makecell{our\\ work}}} & \multirow{3}{*}{No} & - & 2.61 & 7.36 & 2.77 & 7.88\T\\
\cline{3-7}
& & GST & 2.57 & 7.43 & 2.72 & 7.82\T\\
\cline{3-7}
& & i-vector & 2.63 & 7.63 & 2.79 & 7.89\T\\
\cline{2-7}
& \multirow{3}{*}{Yes} &- & 2.43 & 7.03 & 2.66 & 7.37\T\\
\cline{3-7}
& & GST & 2.35 & 7.05 & \textbf{2.50} & 7.29\T\\
\cline{3-7}
& & i-vector& \textbf{2.32} & \textbf{6.72} & 2.53 & \textbf{7.19}\T\\
\hline
\hline
\multirow{2}{*}{\cite{DBLP:journals/corr/abs-1811-00707}} & \multirow{2}{*}{-} & - & & & 5.10 & 16.21\T\\
\cline{3-7}
& & GST & & & 4.66& 15.47\T\\
\hline
\end{tabular}
\end{center}
\vspace{-1.2em}
\end{table}

\section{Conclusion}

We presented a straight-forward approach to generate and add synthetic audio data to state-of-the-art end-to-end ASR systems. We showed that we can improve a strong low-resource baseline system that already uses data augmentation and an additional language model by up to 33\% in relative WER on LibriSpeech test-clean, and by 9\% on test-other. The improvements by using synthetic data were larger when used together with SpecAugment and LM combination. Our TTS system uses global-style-tokens for unsupervised speaker embeddings, thus removing the need for speaker labeled training data. By using Griffin \& Lim synthesis as vocoder approach, synthesizing data is computationally inexpensive compared to the ASR training itself. Although we observed large improvements when using TTS data, manual evaluation revealed that the TTS outputs are still poor in stability and speaker adaptation capabilities. Preliminary results on the full \textit{LibriSpeech-960h} corpus show only minor improvements. In future work, we will try to build stronger and more stable TTS systems that include all of the LibriSpeech data as well as investigating possible underfitting that occurs in a large resource environment.

\section{Acknowledgements}
\footnotesize
\begin{wrapfigure}[5]{l}{0.16\textwidth}
	\vspace{-2.4em}
    \begin{center}
        \includegraphics[width=0.18\textwidth]{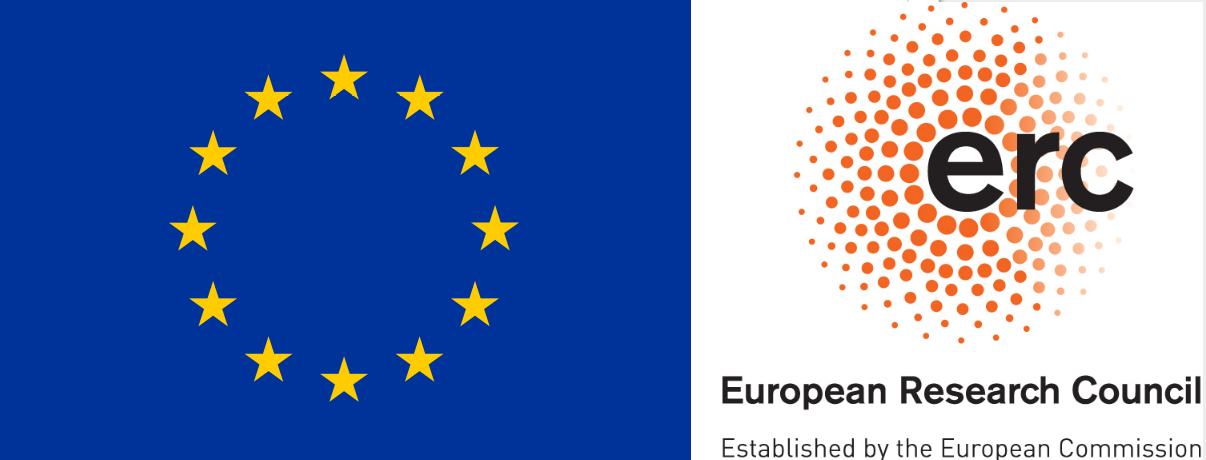}
    \end{center}
\end{wrapfigure}
This work has received funding from the European Research Council (ERC) under the European Union’s Horizon 2020 research
and innovation programme (grant agreement No 694537, project ”SEQCLAS”) and from a Google Focused Award. The work reflects only the authors’ views and none of
the funding parties is responsible for any use that may be made of the information it contains. Simulations were partially performed with computing resources granted by RWTH Aachen University under project nova0003.

\small
%


\bibliographystyle{IEEEbib}

\SetTracking{encoding=*}{-15}\lsstyle  

\let\OLDthebibliography\thebibliography
\renewcommand\thebibliography[1]{
  \OLDthebibliography{#1}
  \setlength{\parskip}{0pt}
  \setlength{\itemsep}{0pt plus 0.07ex}
}

\bibliography{refs}

\end{document}